\documentclass[11pt, a4paper]{thuc3i}
\usepackage[compress]{natbib}
\setheadertext{AutoSFT}

\usepackage{amsmath,amsfonts,bm}

\def\eqref#1{equation~\ref{#1}}

\def\1{\bm{1}}

\DeclareMathAlphabet{\mathsfit}{\encodingdefault}{\sfdefault}{m}{sl}
\SetMathAlphabet{\mathsfit}{bold}{\encodingdefault}{\sfdefault}{bx}{n}

\usepackage{booktabs}
\usepackage{tabularx}
\usepackage{multirow}
\usepackage[utf8]{inputenc} %
\usepackage[T1]{fontenc}    %
\usepackage{hyperref}       %
\usepackage{url}            %
\usepackage{tabularx}
\usepackage{makecell}
\usepackage{array,ragged2e,booktabs}
\newcolumntype{P}[1]{>{\RaggedRight\arraybackslash}p{#1}}
\usepackage{longtable}
\usepackage{amsfonts}       %
\usepackage{amssymb}        %
\usepackage{amsthm}         %

\usepackage{nicefrac}       %
\usepackage{microtype}      %
\usepackage{xcolor}         %

\usepackage{algorithm}
\usepackage{algorithmic}

\definecolor{darkblue}{rgb}{0, 0, 0.5}
\hypersetup{colorlinks=true, citecolor=darkblue, linkcolor=darkblue, urlcolor=darkblue}

\usepackage{xurl}

\usepackage{soul}
\usepackage{amsmath}
\usepackage{subcaption}
\usepackage{graphicx}
\usepackage{changes}
\usepackage{mathtools}
\usepackage{pifont}
\usepackage{tocloft}

\usepackage{colortbl}
\usepackage{wrapfig}
\usepackage{xspace}
\usepackage{multirow}
\usepackage{enumitem}

\usepackage{listings}
\usepackage{fontawesome5}

\usepackage{caption}

\usepackage{afterpage}
\usepackage{float}
\usepackage[edges]{forest}
\usepackage[normalem]{ulem}
\usepackage{bbding}
\usepackage[most,skins,theorems]{tcolorbox}
\usepackage{epigraph}

\usepackage[tikz]{bclogo}
\usepackage[framemethod=tikz]{mdframed}

\definecolor{lightorange}{HTML}{faa755}
\definecolor{lightblue}{RGB}{220,235,250}
\tcbset{
  takeawaysbox/.style={
    title=Takeaways,
    colback=lightblue!80,
    colframe=black,
    fonttitle=\bfseries\small,
    coltitle=white,
    colbacktitle=black,
    enhanced,
    attach boxed title to top left={xshift=2.5mm,yshift=-2.5mm},
    boxed title style={rounded corners, size=small, colframe=black, colback=black},
    width=\linewidth,
    arc=3.5mm
  },
  examplebox/.style={
    colback=orange!6,
    colframe=lightorange,
    fonttitle=\bfseries\small,
    coltitle=black,
    colbacktitle=orange!18,
    enhanced,
    attach boxed title to top left={xshift=2.5mm,yshift=-2.5mm},
    boxed title style={rounded corners, size=small, colframe=lightorange, colback=orange!18},
    width=\linewidth,
    arc=3.5mm
  }
}
\tcbset{
  taskbox/.style={
    title=Task Definition,
    colback=lightblue!80,
    colframe=black,
    fonttitle=\bfseries\small,
    coltitle=white,
    colbacktitle=black,
    enhanced,
    attach boxed title to top left={xshift=2.5mm,yshift=-2.5mm},
    boxed title style={rounded corners, size=small, colframe=black, colback=black},
    width=\linewidth,
    arc=2mm
  }
}

\definecolor{overviewbg}{HTML}{FFF3E0}
\definecolor{overviewframe}{HTML}{E65100}
\tcbset{
  overviewbox/.style={
    title=Overview,
    colback=overviewbg,
    colframe=overviewframe,
    fonttitle=\bfseries\small,
    coltitle=white,
    colbacktitle=overviewframe,
    enhanced,
    attach boxed title to top left={xshift=2.5mm,yshift=-2.5mm},
    boxed title style={rounded corners, size=small, colframe=overviewframe, colback=overviewframe},
    width=\linewidth,
    arc=3.5mm
  }
}

\definecolor{tplhead}{HTML}{1F2A6E}   
\definecolor{tplbodybg}{HTML}{EEEFF9} 
\definecolor{tplkw}{HTML}{1F5FBF}     
\definecolor{tplcmt}{HTML}{8A8A8A}    
\definecolor{tplnum}{HTML}{3A6EA5}    
\newtcolorbox{tplbox}[1]{%
  enhanced, colback=white, colframe=tplhead, colbacktitle=tplhead,
  coltitle=white, fonttitle=\bfseries\footnotesize, title={#1},
  boxrule=0.9pt, titlerule=0pt, arc=2pt,
  left=5pt, right=5pt, top=3pt, bottom=3pt,
  equal height group=tplex,   
}
\newtcolorbox{tplinput}{%
  colback=tplbodybg, colframe=tplbodybg, boxrule=0pt, arc=1.5pt,
  left=4pt, right=4pt, top=2pt, bottom=2pt,
  before skip=3pt, after skip=3pt,   
}

\definecolor{tpltokctrl}{HTML}{1F5FBF}   
\definecolor{tpltokrole}{HTML}{9B2D86}   
\definecolor{tpltokthink}{HTML}{1E8449}  
\definecolor{tpltokesc}{HTML}{8A8A8A}    

\lstdefinestyle{tplpy}{%
  language=Python, basicstyle=\ttfamily\scriptsize,
  keywordstyle=\color{tplkw}\bfseries, commentstyle=\color{tplcmt},
  numbers=left, numberstyle=\tiny\color{tplnum}, numbersep=5pt,
  showstringspaces=false, breaklines=true, frame=none,
  aboveskip=2pt, belowskip=2pt, xleftmargin=12pt,
}

\definecolor{phenobg}{HTML}{EBF5FB}
\definecolor{phenoframe}{HTML}{2980B9}
\definecolor{mechbg}{HTML}{FEF5E7}
\definecolor{mechframe}{HTML}{E67E22}
\definecolor{recipebg}{HTML}{D5F5E3}
\definecolor{recipeframe}{HTML}{1E8449}
\definecolor{overviewdark}{HTML}{2C3E50}

\usepackage{cleveref}

\usepackage[dvipsnames]{xcolor}

\tcbuselibrary{listings,breakable,xparse,skins}
\definecolor{tracebg}{HTML}{faf8f3}
\definecolor{traceaccent}{HTML}{a66b4f}
\definecolor{tracenote}{HTML}{6b655a}
\definecolor{tracewarn}{HTML}{9c3b2e}
\DeclareTCBListing{tracebox}{O{}m!O{}}{%
  breakable=false, enhanced, listing engine=listings, listing only,
  listing options={basicstyle=\scriptsize\ttfamily, breaklines=true,
    breakatwhitespace=false, columns=fullflexible, keepspaces=true,
    escapeinside={|}{|}, xleftmargin=0pt},
  boxsep=0pt, left=10pt, right=10pt, top=3pt, bottom=3pt, arc=1pt,
  boxrule=0.5pt, leftrule=3pt, rightrule=0.5pt, toprule=0.5pt,
  bottomrule=0.5pt, colframe=traceaccent!50, colback=tracebg,
  borderline west={3pt}{0pt}{traceaccent}, #3}

\definecolor{skillbg}{HTML}{f4f6f9}
\definecolor{skillaccent}{HTML}{3f6187}
\definecolor{skillkey}{HTML}{2f4f6f}
\DeclareTCBListing{skillbox}{O{}m!O{}}{%
  breakable=false, enhanced, listing engine=listings, listing only,
  listing options={basicstyle=\scriptsize\ttfamily, breaklines=true,
    breakatwhitespace=false, columns=fullflexible, keepspaces=true,
    escapeinside={|}{|}, xleftmargin=0pt},
  boxsep=0pt, left=10pt, right=10pt, top=3pt, bottom=3pt, arc=1pt,
  boxrule=0.5pt, leftrule=3pt, rightrule=0.5pt, toprule=0.5pt,
  bottomrule=0.5pt, colframe=skillaccent!50, colback=skillbg,
  borderline west={3pt}{0pt}{skillaccent}, #3}

\definecolor{casebg}{HTML}{fbfbfa}
\definecolor{caseaccent}{HTML}{55606e}
\definecolor{runbase}{HTML}{b4473f}
\definecolor{runplus}{HTML}{2f5d9e}
\newtcolorbox{casebox}[1][]{%
  breakable=false, enhanced, boxsep=0pt, left=8pt, right=8pt, top=5pt,
  bottom=5pt, arc=1pt, boxrule=0.5pt, leftrule=3pt, rightrule=0.5pt,
  toprule=0.5pt, bottomrule=0.5pt, colframe=caseaccent!40, colback=casebg,
  borderline west={3pt}{0pt}{caseaccent}, fontupper=\small, #1}

\NewDocumentCommand{\cheng}
{ mO{} }{\textcolor{orange}{\textsuperscript{\textit{Cheng}}\textsf{\textbf{\small[#1]}}}}

\title{AutoSFT: An Agentic System for Automated Data Research in Supervised Fine-Tuning}

\author{%
    Bingxiang He$^{*\dagger1}$,
    Pengqi Sun$^{*2}$,
    Shihao Peng$^{*1}$,
    Shaoxuan Li$^{*1}$,
    Huiyu Xiao$^{*2}$,
    Hongjian Shi$^{*3}$,
    Zixuan Fu$^{*1}$,
    Ziqing Qiao$^{*1}$,
    Yuxin Zuo$^{*\dagger1}$,
    Cheng Qian$^{4}$,
    Xinqiang Yu$^{1}$,
    Jiayi Li$^{3}$,
    Xiusi Chen$^{5}$,
    Yizhen Song$^{3}$,
    Xiangyang Wang$^{1}$,
    Iofai Lei$^{1}$,
    Yudong Wang$^{1}$,
    Ning Ding$^{1}$,
    Zhiyuan Liu$^{1}$,
    Xiaozhi Wang$^{\ddagger1}$,
    Chaojun Xiao$^{\ddagger1}$\\
    $^{1}$Tsinghua University \quad
    $^{2}$Beijing University of Posts and Telecommunications \quad
    $^{3}$ModelBest Inc.\\
    $^{4}$University of Illinois Urbana-Champaign \quad
    $^{5}$The Chinese University of Hong Kong \quad
    \vskip -1mm
    \textbf{$^\dagger$Project Lead.} \quad \textbf{$^*$Core Contributors.} \quad\textbf{$^\ddagger$Corresponding Authors.}
    \vskip -1mm
    \faGithub~{\fontfamily{ppl}\selectfont \textbf{Code:}} ~\url{https://github.com/Thinking-Space/AutoSFT}. \\
    \vskip 1mm
    \faEnvelope[regular] \texttt{hebx24@mails.tsinghua.edu.cn, xzwang@sz.tsinghua.edu.cn, xcj@tsinghua.edu.cn}
}

\begin{document}

\begin{abstract}
Language model agents increasingly automate post-training, but when given the full training stack they converge on supervised fine-tuning and spend most effort revising training data.
Existing systems give the agent one kind of control, selecting rows from a fixed pool or calling a generator, whereas a human data engineer adjusts proportions, prunes, repairs, replaces and synthesizes the data.
We design a data action space for these changes and ask whether an automated system can make the full range rather than one of them, and what structure that requires.
We first run this loop with a single agent, and it reaches only one action, adjusting category proportions, while its context fills with execution output and is discarded mid-search.
We therefore introduce \textbf{AutoSFT}, a multi-agent system where a meta agent separates execution from analysis, gathers proposals from independent angles, and calls a synthesis subagent for capabilities the pool cannot cover, with the model, the trainer, and the budget fixed within each comparison.
Adding each component in turn helps: the executor halves context loss, the analysis subagents lift the best IFEval by $+6.5$, and synthesis adds $+3.4$ IFEval and $+2.9$ IFBench. The same loop transfers to code, where analysis adds $+3.7$ MBPP+ over a single agent that does not improve.
Every result comes from runs that retrain from the base model each round, so each component's effect is individually attributable.
Our findings suggest that automating SFT data research depends more on covering diverse data actions than on a stronger single action.
\end{abstract}

\maketitle

\section{Introduction}
\label{sec:introduction}

LLM agents now run long workflows with little human oversight, and several systems use them to automate parts of AI research itself~\citep{aiscientist2024,airesearcher2025,adas2025,anthropicaar2026}.
What makes such a loop searchable is that every proposal can be evaluated automatically, and post-training has this property, so recent benchmarks evaluate agents on it~\citep{posttrainbench2026,agent2rlbench2026}.
Given control of the whole training stack, agents converge on one part: across the runs audited by PostTrainBench~\citep{posttrainbench2026}, they adopt supervised fine-tuning (SFT) and spend most of their remaining effort revising the training data.
This agrees with a broader finding that a carefully curated dataset can match or outperform a much larger one~\citep{lima2023,chen2024minprompt,s12025}, and that data selection and arrangement affect downstream performance directly~\citep{righttime2025,delift2024}.
Recent systems therefore treat the data as the main search object~\citep{autoselection2026,agenticde2026,andes2026,autodata2026}.

Each existing system gives the agent a single kind of control, selecting rows from a fixed pool or calling a generator.
We ask \textit{whether an automated agent can instead command the full space of operations that an experienced data engineer uses.}
Such an engineer adjusts category proportions, prunes and repairs rows, replaces overlong responses, writes examples for missing capabilities, and orders the data.
We write these six kinds of change as a data action space (\Cref{sec:setup}), and use it only after a run to measure how much of that space the agent reaches.

We first study this question with a single agent, restarting each round from the same base checkpoint.
The agent collapses onto one action family, repeats related recipes, and spends most of its context on raw command output rather than data analysis.
\Cref{sec:prelim} examines these failures, the trace evidence, and our fix, explaining why an undivided research loop is hard to sustain.
One agent analyzes behavior, proposes data changes, executes experiments, and interprets logs, so execution competes with analysis for context and a single proposal stream reinforces recent choices.
This yields two requirements: isolate execution from data analysis, and maintain complementary views of the same model behavior during proposal generation.

\begin{figure}[t]
    \centering
    \includegraphics[width=0.9\linewidth]{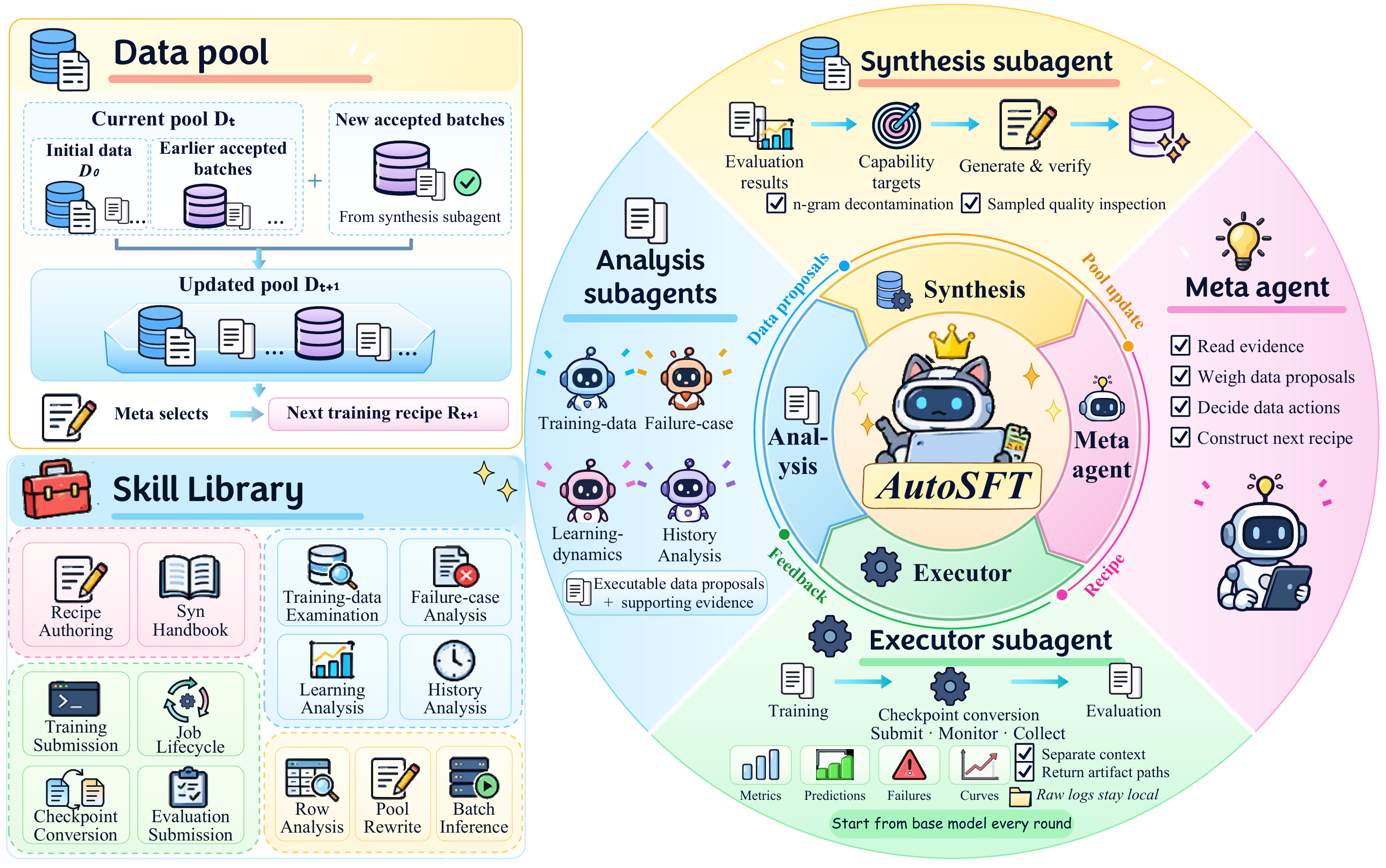}
    \caption{Overview of AutoSFT. The meta agent commits a recipe. An executor trains and evaluates it separately and returns artifact paths. Analysis subagents read these artifacts from complementary angles, and a synthesis subagent writes data for undersupplied capabilities. The meta agent integrates their proposals into the next recipe. All roles share a candidate data pool and a skill library.}
    \label{fig:method:overview}
\end{figure}

AutoSFT (\Cref{sec:method}) is a multi-agent system built around these two requirements.
A meta agent owns each round and delegates to three kinds of subagent.
An executor runs training, conversion, and evaluation separately, returning file paths and a short status report so the meta agent can spend its context on data.
Analysis subagents read the same round from four angles: training data, failures, training signals, and earlier rounds, each returning one proposed data action.
A synthesis subagent writes new data for capabilities the pool barely covers, which neither of the first two can add.

We add the three components one at a time, fixing the base model, initial pool, token budget, training script, and evaluation within each comparison, so any difference is attributable to the added component (\Cref{sec:exp}).
The executor halves context loss, the analysis subagents broaden the range of changes and raise best IFEval by $+6.5$, and synthesis adds $+3.4$ IFEval and $+2.9$ IFBench by covering undersupplied capabilities.
The same loop transfers to code domain without a ready-made capability taxonomy, adding $+3.7$ MBPP+ over a single agent that does not improve (\Cref{sec:generalization}).
Together, these results suggest that automating SFT data research depends less on a stronger single action than on structure that lets an agent cover the range of changes a data engineer makes.

\noindent\textbf{Contributions.} Our contributions are threefold.
\begin{itemize}[topsep=0pt, partopsep=0pt, leftmargin=12pt, itemsep=0pt]
\item We frame automated SFT data research around a space of six data-engineering actions, and show that a corrected single-agent loop reaches only one of them (\Cref{sec:setup,sec:prelim}).
\item We build \textbf{AutoSFT}, a multi-agent system that isolates execution, gathers proposals from complementary analysis angles, and synthesizes data for capabilities the pool lacks (\Cref{sec:method}).
\item Adding each component while retraining from the base model, so its effect is attributable, widens the actions a run takes and improves the recipe, by $+6.5$ IFEval from analysis, $+3.4$ IFEval and $+2.9$ IFBench from synthesis, and $+3.7$ MBPP+ in a code domain (\Cref{sec:exp}).
\end{itemize}

\section{Automated SFT Data Research}
\label{sec:setup}

\subsection{Research loop}
\label{sec:setup:loop}

\begin{figure}[t]
    \centering
    \includegraphics[width=0.8\linewidth]{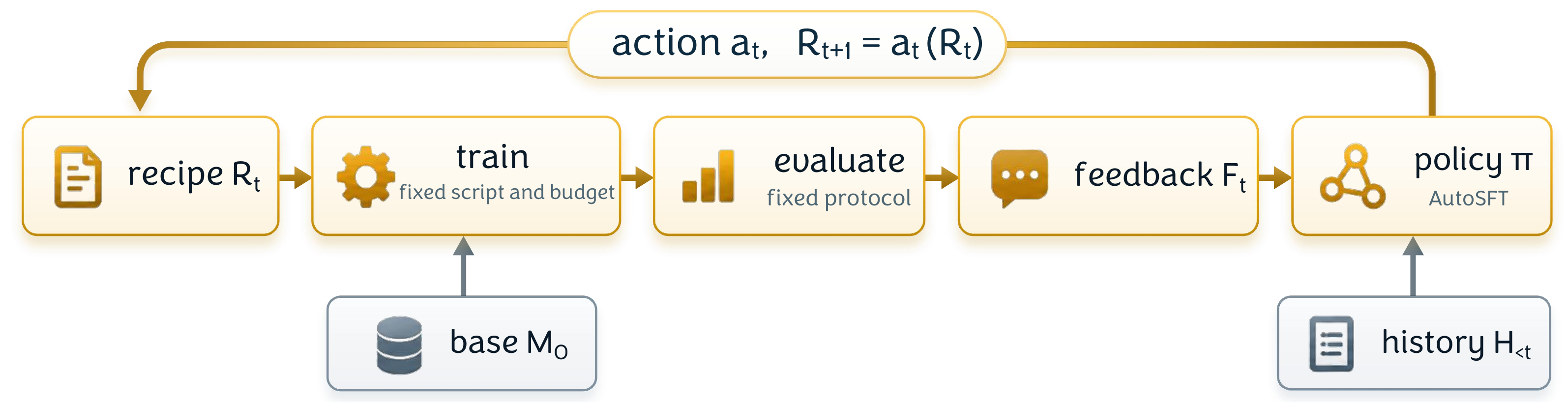}
    \caption{One round of the research loop. Only $\pi$ varies across the configurations we compare.}
    \label{fig:setup:loop}
\end{figure}

An automated SFT data research loop searches over the data recipe that reaches the trainer while the base checkpoint stays fixed, as shown in \Cref{fig:setup:loop}.
We fix a base checkpoint $M_0$ and an initial candidate pool $D_0$.
A recipe $R_t$ gives every example in the round's pool $D_t$ a sampling weight, possibly zero, and any rows written during round $t$ join the pool, so that $D_{t+1}$ extends $D_t$.
Training always re-initializes from $M_0$, so $\mathrm{train}(M_0, R_t) \to C_t$ and the pool and the recipe are the only quantities that carry research history across rounds.

We evaluate $C_t$ on benchmarks $\{T_k\}_{k=1}^{K}$ under one fixed protocol.
A round returns the benchmark scores together with the per-example failures, the training dynamics, and a comparison against what earlier rounds tried, and we write this bundle as the feedback $F_t$.
The recipe then advances through
\[
a_t = \pi(D_t, R_t, F_t, H_{<t}), \qquad R_{t+1} = a_t(R_t),
\]
where $a_t$ is a data action from the space $\mathcal{A}$ of \Cref{sec:setup:actions}, $R_{t+1}$ is a recipe over $D_{t+1}$, and $H_{<t}$ holds the recipes, feedback, and actions of every earlier round.
We call $\pi$ the research policy which is the procedure under study, turning a round's feedback and the run's history into the next data action.
AutoSFT is one instance of research policy, implemented as a multi-agent system.

The trainer is a single SFT script that runs three epochs, and every recipe within a comparison is capped at the same token budget.
Any two rounds therefore differ only in which rows they trained on, and no agent can trade training data for extra rounds.
The loop runs until an objective is achieved.
\Cref{sec:prelim} reports the evidence behind the from-base rule, the fixed budget, and the score target.

\subsection{Data research actions}
\label{sec:setup:actions}

A recipe is built out of rows, where a row carries an instruction, a response, and the metadata that describes it.
An action states which rows to train on, how often each appears, in what order, and whether any row is edited or newly written.
Agents write recipes in free form with no field schema, operator list, and statement that a particular kind of change is available.

\begin{figure}[t]
    \centering
    \includegraphics[width=\linewidth]{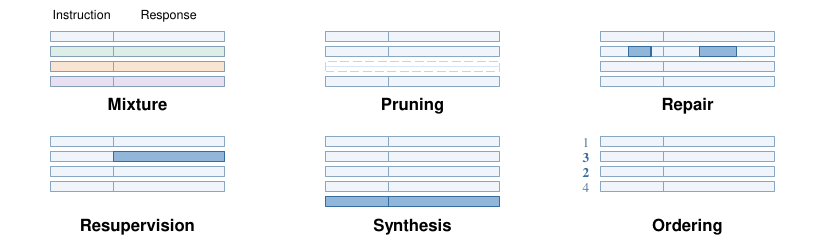}
    \caption{The six families of data action, drawn as their effect on the rows of a recipe. Shading marks what the action changes.}
    \label{fig:setup:actions}
\end{figure}

\Cref{fig:setup:actions} describes the six families of data action a data researcher makes most often.
Each family is fixed by what happens to the rows and by the level at which the change is decided, which ranges from a whole category through a single row and a capability to the recipe as a whole, so an action can be labelled from the recipe diff without reading the agent's stated reason.

Every family is reachable in every configuration we run, since an agent that can write a recipe can also call a model to rewrite a response or produce a new row.
We apply the labels only after a run and report how many families a run covered alongside its scores.

Actions that leave the recipe unchanged, such as holding out a per-category validation set or evaluating a checkpoint more times, are research actions rather than data actions and stay outside $\mathcal{A}$.

\section{Pilot study: What limits single agents in SFT data research}
\label{sec:prelim}

We study how a single agent behaves when it owns an entire SFT data research loop.
We reproduce the open-ended PostTrainBench setting~\citep{posttrainbench2026} for Qwen3-1.7B-Base~\citep{yang2025qwen3} on IFEval~\citep{zhou2023instructionfollowingevaluationlargelanguage}, in which the agent receives one H100, unrestricted internet access, and a ten-hour wall-clock budget.
Six design choices in this setting interfere with measuring what the agent does with data, and we correct them before running the loop: we fix the training algorithm to SFT, provide an initial dataset pool instead of Internet search, add a second benchmark IFBench~\citep{pyatkin2025generalizing}, retrain from the base checkpoint each round, replace the wall-clock allowance with benchmark targets, and return richer feedback including training loss, category-level scores, and per-example failures.
Details are deferred to Appendix~\ref{app:prelim}.

\subsection{Three failure modes of the single-agent loop}
\label{sec:prelim:loop}

\noindent\textbf{Setup.}
The candidate pool is argilla/ifeval-like-data\footnote{\url{https://huggingface.co/datasets/argilla/ifeval-like-data}}, a verifiable instruction-following corpus~\citep{magpie2024}.
Each round draws a recipe of at most 10M tokens and trains for three epochs at sequence length 32{,}768, with global batch size 16 and learning rate $10^{-4}$.
The loop stops when IFEval reaches 80 and IFBench reaches 50.
A single Codex agent running GPT-5.5 at medium reasoning effort submitted 33 rounds and completed evaluation on 32 of them within 11.5 hours, at which point neither target had been reached.

\begin{figure*}[!t]
    \centering
    \begin{minipage}[c]{0.75\linewidth}
        \centering
        \includegraphics[width=\linewidth]{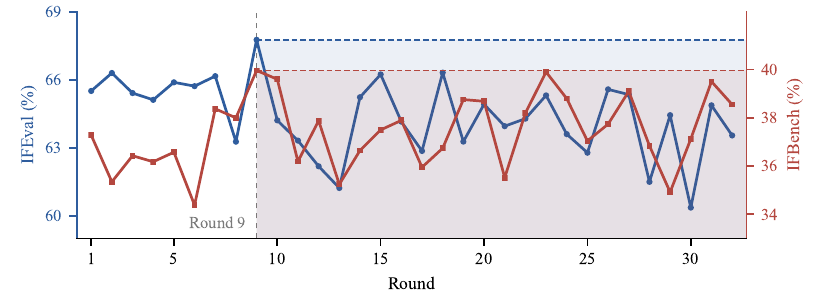}
    \end{minipage}
    \hfill
    \begin{minipage}[c]{0.24\linewidth}
        \centering
        \includegraphics[width=\linewidth]{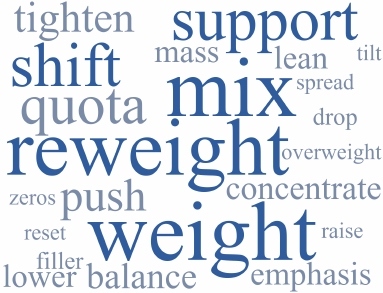}
    \end{minipage}
    \caption{\textbf{Left}: benchmark scores over the 32 evaluated rounds on separate axes. \textbf{Right}: the terms the agent used to state its chosen data action, sized by how often it used them.}
    \label{fig:prelim:trace}
\end{figure*}

\noindent\textbf{The search stays inside one family of actions.}
Every round adjusts how much of each category to train on, so the whole run falls in the mixture family.
As \Cref{fig:prelim:trace} shows, round 9 is the best on both benchmarks and the 23 rounds after it produce nothing better.
The agent does not lack terms for the move it repeats, using 22 distinct expressions across the run, from \emph{reweight} to \emph{raise} and \emph{concentrate}.
What it lacks is variety in the move itself.
Its reasoning never mentions repairing a row, replacing a response, or writing a new example, and the corresponding terms do not appear in the trace.

\begin{wrapfigure}{r}{0.5\textwidth}
\vspace{-10pt}
\centering
\includegraphics[width=\linewidth]{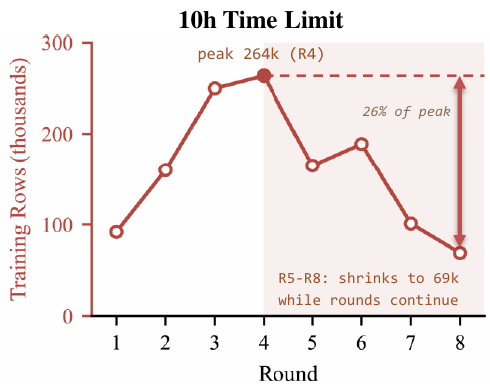}
\vspace{-10pt}
\caption{Training data per round under a wall-clock budget.}
    \label{fig:prelim:objective}
\vspace{-20pt}
\end{wrapfigure}

\noindent\textbf{Time Limits Reduce Per-Round Training Data.}
The open-ended run is allotted ten hours, and \Cref{fig:prelim:objective} reports the amount of data presented to the trainer in each round.
The recipe expands over the first four rounds, reaching 264k rows, and then declines in every subsequent round, ending at 69k rows, which is 26\% of the peak.
The agent continues to submit rounds throughout, so the run does not terminate early.
Rather, it progressively reduces the cost of each remaining round.
Once a checkpoint is sufficiently good to retain, additional training data yields no further improvement under the objective, whereas the time it would cost can be used for another attempt.
Under benchmark targets, this contraction does not occur: recipe size is around max token budget, and the agent never refers to the remaining time budget.

\begin{figure}[t]
    \centering
    \includegraphics[width=\linewidth]{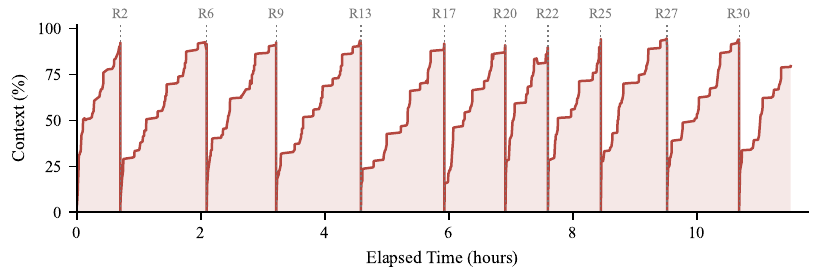}
    \caption{The single agent's context. Dashed lines mark compactions labelled with the round.}
    \label{fig:prelim:context}
\end{figure}

\noindent\textbf{The context fills and is discarded ten times.}
The agent runs data decisions, job submission, checkpoint conversion, and evaluation in one session, and \Cref{fig:prelim:context} shows that session filling to 94.7\% of a 258{,}400-token window and being compacted ten times, first during round 2 and then every three to four rounds.
Research history is therefore discarded and rebuilt ten times while the search is still running.
The context is not dominated by the obvious suspect.
Polling job status is the most frequent operation, at 371 calls, yet it accounts for only 1.7\% of the 9.9M bytes accumulated.
The space goes instead to a few large payloads: training output at 13.4\%, repeated reads of the run record at 11.8\%, applying the data action at 9.7\%, and evaluation output at 8.9\%. With the agent's reasoning (16.3\%) and compaction summaries (8.3\%), six sources take two thirds of the window.
None of them relates to choosing the next data action, which argues for giving those steps their own session.

\subsection{From observations to requirements}
\label{sec:prelim:requirements}

The second observation is answered by the objective itself, and the controlled loop already carries that change.
The other two set the structure AutoSFT has to provide.
Execution has to leave the deciding agent's context, so a dedicated executor runs submission, conversion, and evaluation in its own session and returns status receipts and artifact paths (\Cref{sec:method:executor}).
Proposals have to come from multiple lines of reasoning, so we run several analysis subagents from different angles, following \citet{anthropicaar2026}, who find that distinct starting directions preserve diversity across parallel agents (\Cref{sec:method:analysts}).
Besides, neither change introduces data that the pool did not already contain, which is a separate limitation and the reason \Cref{sec:method:synthesis} introduces a third component.
\section{AutoSFT}
\label{sec:method}

AutoSFT instantiates the research policy $\pi$ of \Cref{sec:setup:loop} and shares $M_0$, $D_0$, and the training and evaluation protocol with the single agent of \Cref{sec:prelim}.
A meta agent owns the round and delegates to three kinds of subagents: \textbf{an executor subagent}, \textbf{four analysis subagents}, and \textbf{a synthesis subagent}, whose information flow is shown in \Cref{fig:method:overview}.
The meta agent hands $D_t$ and $R_t$ to the executor, which trains and evaluates in a separate session and returns the feedback $F_t$.
The analysis and synthesis subagents then consult the relevant parts of $D_t$, $R_t$, $F_t$, and $H_{<t}$ through their associated files. Each analysis subagent proposes a change to the recipe from its own perspective, while the synthesis subagent generates data for undersupplied capability targets to form $D_{t+1}$.
The meta agent weighs these proposals against its own reading and settles on $a_t$, giving $R_{t+1} = a_t(D_{t+1}, R_t)$. Implementation details, including a skill library and memory design, are given in Appendix~\ref{app:method}.

\subsection{Executor subagent}
\label{sec:method:executor}

As shown in \Cref{sec:prelim:loop}, the context of the single-agent loop fills with the verbose logs and outputs of training and evaluation jobs, leaving the information that matters unused. 
Given this, we introduce an executor, which takes the entire lifecycle of training, checkpoint conversion, and evaluation out of the meta agent and maintains a context of its own.
Specifically, everything it does is one cycle of \textit{submitting a job, monitoring it to completion, and collecting its artifacts}: the cycle runs once on the training job to obtain the round's model $C_t$, and then once on each of the $K$ benchmarks to obtain the evaluation results for $C_t$, with the raw logs and intermediate outputs staying in its own context.
When the workflow finishes, the executor saves the feedback artifacts and a structured execution report to files and returns only their paths. The meta agent and the analysis and synthesis subagents can then read the relevant
files as needed, without receiving the full execution logs.

\subsection{Analysis subagents}
\label{sec:method:analysts}

\Cref{sec:prelim:loop} shows that all 33 rounds of the single agent stay inside the mixture family: it pursues the objective through one kind of strategy.
To break out of the limitation, we introduce four analysis agents, each of which examines the current round from a different evidence source: the training data, failure cases, learning dynamics, or research history.

The four subagents examine each round using complementary evidence.
\textbf{The training-data subagent} reads $D_t$ and $R_t$, assesses the selected data for quality, difficulty, coverage, and redundancy, and proposes changes to its content or composition under the fixed token budget. 
\textbf{The failure-case subagent} compares successful and failed cases in $F_t$ to identify recurring failure patterns and propose supervision to address them.
\textbf{The learning-dynamics subagent} analyzes the training loss and other monitoring signals in $F_t$ in relation to the data used for training, seeking evidence that explains stalled learning progress. 
\textbf{The history subagent} reads $H_{<t}$, distinguishes failed interventions from ideas that have not been adequately tested, and proposes corrections or renewed trials. 
Each subagent returns one executable data proposal with supporting evidence.

\subsection{Synthesis subagent}
\label{sec:method:synthesis}
In practice, the initial pool $D_0$ may not cover every capability the benchmarks require, and for others it holds examples but too few to constitute effective supervision.
We therefore introduce a synthesis subagent to generate examples for undersupplied capability targets.
Such a target is convenient to state and to verify in the instruction-following setting: it is a single checkable constraint, or a combination of constraints that co-occur, such as \textit{repeating the instruction verbatim while using no comma}, and in either case a programmatic verifier decides whether an example satisfies it.

Concretely, the synthesis agent is invoked in every round, reads the evaluation results, and derives a set of targets to strengthen.
For each target whose supply in $D_t$ is insufficient, it synthesizes examples that pass the target's verifier.
If no target is undersupplied in a given round, the synthesis agent generates no data.
Generated data pass verification of the target's constraint, an independent semantic review, and an $n$-gram overlap check against the evaluation sets, described in Appendix~\ref{app:method:synthesis-validation}.
Those that pass enter $D_{t+1}$, and whether the round trains on them remains the meta agent's decision.

\section{Experiment}
\label{sec:exp}

AutoSFT adds three components to the loop of \Cref{sec:prelim}, and we test them one at a time.
Each subsection compares a base configuration against the same configuration with one component added, so a difference is attributable to that component alone, while \Cref{sec:generalization} runs the full system on a coding domain to test whether the same structure transfers.
Appendix~\ref{app:implementation-cost} gives the full implementation settings for every comparison and an estimate of the compute and API cost of a run.

\subsection{Executor subagent}
\label{sec:exp:executor}

\noindent\textbf{Setup.}
The base configuration is the single agent of \Cref{sec:prelim:loop}.
Adding the executor places that lifecycle in a separate session, so the meta agent receives artifact paths and a brief status report rather than executing the lifecycle.
Both runs begin from the same base checkpoint, share the same context window, and are evaluated from snapshots taken at 11.5 hours.
All other settings are identical.

\begin{figure}[t]
    \centering
    \includegraphics[width=\linewidth]{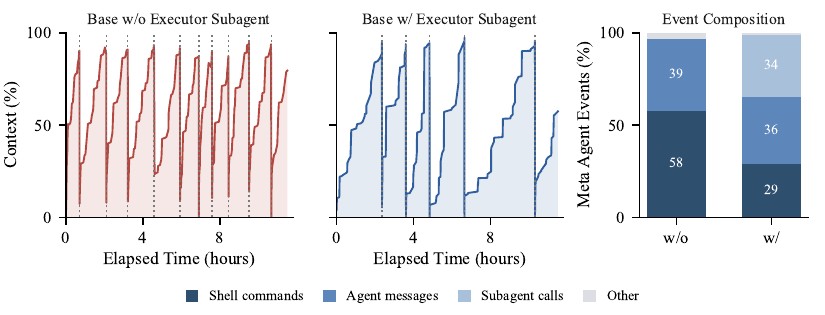}
    \caption{Context used by the meta agent over each run and the composition of its recorded events.}
    \label{fig:exp:executor}
\end{figure}

\noindent\textbf{Result.}
The executor halves how often the meta agent loses its context.
Over comparable elapsed time the base run compacts ten times and the executor run five (\Cref{fig:exp:executor}).
The composition of the meta agent's work shows where the pressure was coming from.
Shell commands fall from 58\% of its recorded events to 29\%, and a third of what remains is delegation to the executor.
Status checks are the clearest instance of the material that leaves.
The base agent issues 268 of them and spends 23.0\% of its commands asking whether a job has completed.
The base run rebuilds a context dominated by command output, whereas the executor run rebuilds a context containing the round's analysis.

\subsection{Analysis subagent}
\label{sec:exp:ablations}

\noindent\textbf{Setup.}
The base configuration uses one agent to read each round's artifacts and write the next recipe.
The analysis configuration adds four subagents that read the same round from training data, failed outputs, training signals, and earlier rounds, each proposing one change.
A meta agent makes the final decision. Both configurations use the dataset in \Cref{sec:prelim:loop}, start from the same base checkpoint, train for three epochs under a 3M-token cap, and are evaluated on IFEval using avg@10, per-comparison settings are given in Appendix~\ref{app:implementation-cost}.
Neither reached the target.
We therefore report the base setting over 50 rounds and the analysis setting over 30 rounds under the same time budget.

\begin{figure}[t]
    \centering
    \includegraphics[width=\linewidth]{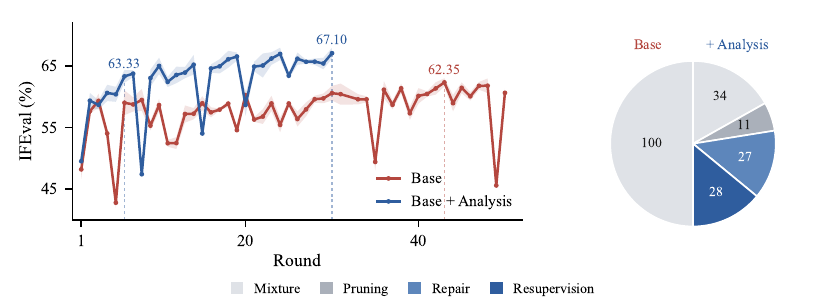}
    \caption{\textbf{Left}: IFEval per round, with dotted guides at the rounds named in the text. \textbf{Right}: distribution of data actions over \Cref{fig:setup:actions} in agent traces for w/ and w/o Analysis settings.}
    \label{fig:exp:analysis}
\end{figure}

\noindent\textbf{Result.}
Analysis improves the score and reaches higher values earlier.
As shown in \Cref{fig:exp:analysis}, the analysis run reaches 63.33 at R6, already above the best score the base run attains in 50 rounds, and climbs to 67.10 at R30.
The base run's best is 60.60 within the first 30 rounds and 62.35 across all 50, so analysis leads by 6.50 points at equal rounds and by 4.75 even when the base run is given 20 rounds more.
The two runs also use the data-action families in different proportions (\Cref{fig:exp:analysis}, right).
With analysis, repair and resupervision together account for 55\% of the actions taken, whereas the base run stays within the mixture family, adjusting category proportions in every round as the single agent of \Cref{sec:prelim:loop} does.
Neither run writes a new example, which is the subject of \Cref{sec:exp:synthesis}, and neither alters the training order because SFT fixes to a shuffling seed.

Furthermore, the actions that analysis unlocks often set a new best score, where a new best denotes a recipe that exceeds every earlier score in the same run.
Selecting data by validation results sets a new best in 41.7\% of its uses, and editing both instructions and answers in 6 of its 14 uses.
The base run never takes these actions, since it stays within the mixture family, so \Cref{tab:app:newbest} in Appendix~\ref{app:exp:analysis} reports its new-best rate only for mixture.
In summary, analysis helps the meta agent read the same evidence in multiple ways, so it can better decide which supervision to change and whether the resulting examples teach the intended behavior.

\subsection{Synthesis subagent}
\label{sec:exp:synthesis}

\noindent\textbf{Setup.}
The base configuration selects from the staged pool and may edit the examples it selects.
Adding the synthesis subagent enables a round to generate new examples for a capability, written against that capability's verifier and admitted to the pool only after passing that verifier.
Both runs train from the same base checkpoint for three epochs under a 100M-token cap that counts generated rows, and every checkpoint is evaluated as avg@10 on IFEval and IFBench.
The budget is larger than in the previous two comparisons because a smaller budget would keep the pool below its size cap, whereas new data becomes valuable only once selection can no longer improve the recipe.

\begin{figure}[t]
    \centering
    \includegraphics[width=\linewidth]{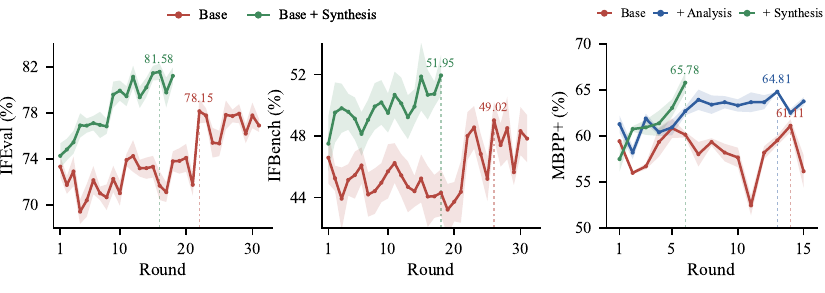}
    \caption{Scores per round. \textbf{Left two panels}: synthesis ablation on instruction following (Base vs Base + Synthesis, \Cref{sec:exp:synthesis}). \textbf{Right panel}: three configurations on code generation (\Cref{sec:generalization}). Bands give the standard deviation over evaluation repeats.}
    \label{fig:exp:synthesis}
\end{figure}

\noindent\textbf{Result.}
Synthesis improves both benchmarks and reaches higher levels earlier.
As shown in \Cref{fig:exp:synthesis} (left, middle), the synthesis run reaches a best IFEval of 81.58, compared with 78.15 for the base run, and a best IFBench of 51.95, compared with 49.02.
Every IFEval checkpoint from R9 onward exceeds the best checkpoint that the base run attains over 31 rounds.
The aggregate gain does not imply that every generated batch helps.
The \texttt{repeat\_prompt} capability remains near 40\% for the first 8 rounds, and the batch supplied at R9 raises it to 75.6\% in the same round, showing a weakness the loop diagnoses and then addresses by generating new data.
By contrast, \texttt{odd\_even\_syllables} is targeted in 3 separate rounds and scores zero in every round, suggesting that supply alone cannot teach a constraint the model cannot yet follow.
In Appendix~\ref{app:synthesis}, we trace the full run of these capabilities, and conduct a comprehensive per-capability comparison across all targets, showing that most capabilities ending higher with synthesis, with the largest gains on repetition, sentence and word counts, and comma avoidance.

\subsection{Generalization to code domain}
\label{sec:generalization}

\noindent\textbf{Setup.} To assess whether AutoSFT remains effective beyond instruction following, we evaluate it on code generation with MBPP+\footnote{\url{https://huggingface.co/datasets/evalplus/mbppplus}}.
Unlike IFEval and IFBench, MBPP+ grades only by test execution and carries no capability breakdown, so the system builds its own capability structure from only the benchmark and the pool.
The backbone is Qwen3-1.7B-Base with MBPP+ target score 70, and the pool holds 742{,}842 Python instruction--code pairs from OPC-SFT-Stage2~\citep{Huang2024OpenCoderTO}, KodCode-v1-SFT-R1~\citep{xu2025kodcode}, and Magicoder-OSS-Instruct-75K~\citep{wei2024magicoder}.
Each round retrains from the backbone for three epochs at sequence length 8{,}192 under a 100M-token budget and is evaluated using avg@3.
We compare these three configurations: Base, Base + Analysis, and Base + Analysis + Synthesis.

\noindent\textbf{Result.}
Within 64 hours, Base and Base + Analysis each complete 15 rounds and Base + Analysis + Synthesis completes 6 (\Cref{fig:exp:synthesis}, right panel).
The single agent shows no sustained improvement, while Base + Analysis produces a consistent upward trend and reaches 64.8\% at round 13.
Adding synthesis incurs additional runtime per round but reaches 65.8\% by round 6.
The structure that helped in instruction following transfers to code: multi-agent analysis lifts the recipe above the single agent, and synthesis reaches a comparable peak in fewer rounds.

\section{Related Work}
\label{sec:related}

\noindent\textbf{Data selection and synthesis for post-training.}
Early works show that curation matters more than scale: a small, well-chosen set can match or outperform a much larger one~\citep{lima2023,s12025}, and data ordering alone shapes generalization~\citep{righttime2025}.
Selection methods prune or reweight examples by information gain or diversity~\citep{delift2024,mig2025}.
Synthesis methods bootstrap instructions from seed sets, evolve complexity, or extract alignment data from aligned models~\citep{selfinstruct2023,wizardlm2023,magpie2024,agentinstruct2024,seif2026}. 
Some data curation methods also optimize specifically for preference data, curating feedback or preference training pairs~\citep{ultrafeedback2024,air2025,bees2025}.
All produce a fixed dataset without closing the loop on downstream training performance.

\noindent\textbf{Automated data optimization.}
A growing body of work attempts to automate data research, closing the loop between data curation and model training.
PostTrainBench~\citep{posttrainbench2026} audits 84 agent runs and finds that almost all converge on SFT with data as the sole search object.
On the selection side, TREX~\citep{trex2026} searches recipes via MCTS, AutoSelection~\citep{autoselection2026} formalizes operator-program search over a fixed pool, DataChef~\citep{datachef2026} trains a policy with a proxy reward, and \citet{agenticde2026} iterates on data alone with gains that plateau after round 8--15.
On the synthesis side, Autodata~\citep{autodata2026} revises generation strategies from solver feedback, ANDES~\citep{andes2026} packages synthesis as a callable skill, and BigBang~\citep{bigbang2026} scales self-evolving synthesis with a critic proxy.
Arbor~\citep{arbor2026}, AutoSOTA~\citep{autosota2026}, and the Anthropic automated researcher~\citep{anthropicaar2026} organize research with hypothesis trees, multi-agent decomposition, and diverse initial directions respectively.
These systems address either selection or synthesis, while we unify both and retrain from base each round so that every action is individually attributable.

\noindent\textbf{Benchmarks for AI research.}
Benchmarks now evaluate agents on ML research tasks at scales from single Kaggle problems to full R\&D pipelines~\citep{mlebench2024,mlagentbench2024,rebench2024,expbench2025,mlrcbench2025,mlsbench2026,researchgym2026,autolab2026}, on AI infrastructure~\citep{kernelbench2025,frontiereng2026,phibench2026,inferencebench2026}, and on self-improvement~\citep{pacebench2026,ai4aibench2026,agent2rlbench2026}.
Closest to our setting, DataEnvGym~\citep{dataenvgym2024} frames data generation as an RL loop, Curation-Bench~\citep{curationbench2026} gives agents iterative control over a fixed pipeline, and RSIBench-Data~\citep{rsibenchdata2026} and DataPrep-Bench~\citep{dataprepbench2026} benchmark data-centric recipe search.
We differ by requiring from-base retraining each round, repeated evaluation to bound noise, and a verification gate on synthesis.

\section{Conclusion}
\label{sec:conclusion}

A single agent owning the full SFT data research loop collapses to one kind of change and loses context repeatedly. 
AutoSFT adds structure that an executor separates execution from analysis, several analysis subagents propose complementary changes, and a synthesis subagent writes data for uncovered capabilities.
Adding each component one at a time under from-base retraining shows that the executor frees the meta agent's context, analysis broadens changes and improves scores earlier, and synthesis reaches capabilities with almost no data in the pool.
The same loop transfers to coding without a ready-made taxonomy, where the system builds its own structure from the benchmark and pool alone.
These results suggest that the bottleneck may not be any single action but the structure that lets an agent cover the full range of data-engineer changes, and that multi-agent decomposition is a practical way to provide it.

\section*{AI use statement}
We used generative AI tools to edit prose for readability, to draft and refine figures, and to assist with code implementation for AutoSFT and the analysis scripts.
We did not use generative AI tools to generate research ideas, design the methodology, or produce any reported result.
The system under study is itself an LLM agent, so the data recipes, analysis, and synthesis prompts it produces are the object of our measurements rather than authored content of this paper.
All reported numbers come from the fixed training and evaluation script rather than from any agent's self-report, and the authors verified every claim against the per-round run ledger.
We have reviewed all AI-assisted work and take responsibility for the final content of this paper.

\section*{Ethics statement}
The instruction-following pool is built from openly released datasets and the coding pool from public instruction--code corpora, cited in \Cref{sec:prelim} and \Cref{sec:generalization}.
Because the AutoSFT optimizes benchmark scores, it could in principle be used to inflate leaderboard numbers, and we report off-target and held-out benchmarks so that a gain from overfitting the target is visible rather than hidden.
We apply an $n$-gram overlap check between generated data and the evaluation sets in Appendix~\ref{app:method:synthesis-validation}, and we report the compute and API cost of the runs in Appendix~\ref{app:implementation-cost}.

\section*{Reproducibility statement}
We will release the AutoSFT implementation, including the training and evaluation script and the agent scaffolding, as anonymized supplementary material.
The fixed training script, the token budget, and the from-base retraining protocol are described in \Cref{sec:setup} and \Cref{sec:prelim}, and the agent roles, prompts, and skill files are given in Appendix~\ref{app:method}.
The initial data pool, the per-round recipes, and the raw evaluation outputs behind every reported number are recorded in a run ledger.

\bibliography{main}

\newpage
\clearpage
\appendix
\section{Details of the single-agent run}
\label{app:prelim}

\subsection{Data action distribution in full trajectory}
\label{app:prelim:trajectory}
\Cref{tab:app:prelim:rounds} lists the action, the realized recipe size, and the two scores for every round.
Round 9 is the best on both benchmarks and no later round improves either one.
Beyond the settings in \Cref{sec:prelim:loop}, all rounds share micro batch size 1 and minimum learning rate $10^{-7}$.

32 rounds are recorded as \textit{reweight} and one as \textit{replay}.
Both are changes of proportion, which is why \Cref{sec:prelim:loop} places the whole run in \emph{mixture}.
Round 24 is the replay, and its records show why the distinction does not survive from-base retraining: it asks to reuse round 9's row selection, but no round inherits anything from an earlier one, so the request is served by reassembling a recipe from the same pool.
The result is 34{,}283 rows against round 9's 34{,}184, with 599 row instances missing and 698 added.
Replaying a category and restoring its proportion produce the same kind of recipe.

\begin{table}[!htbp]
    \caption{Results for all 32 rounds. Action is either reweight (W) or replay (P). The best score on each benchmark is in bold.}
    \label{tab:app:prelim:rounds}
    \centering
    \footnotesize
    \setlength{\tabcolsep}{5pt}
    \begin{tabular}{ccrrr@{\quad\quad\quad}ccrrr}
        \toprule
        \textbf{Round} & \textbf{Action} & \textbf{Rows} & \textbf{IFEval} & \textbf{IFBench} & \textbf{Round} & \textbf{Action} & \textbf{Rows} & \textbf{IFEval} & \textbf{IFBench} \\
        \midrule
        1 & W & 34,860 & 0.6550 & 0.3728 & 17 & W & 32,812 & 0.6286 & 0.3595 \\
2 & W & 34,765 & 0.6630 & 0.3535 & 18 & W & 37,806 & 0.6631 & 0.3674 \\
3 & W & 34,721 & 0.6541 & 0.3643 & 19 & W & 37,433 & 0.6327 & 0.3876 \\
4 & W & 34,431 & 0.6511 & 0.3618 & 20 & W & 37,532 & 0.6492 & 0.3868 \\
5 & W & 32,545 & 0.6588 & 0.3657 & 21 & W & 36,974 & 0.6395 & 0.3552 \\
6 & W & 31,246 & 0.6572 & 0.3438 & 22 & W & 37,342 & 0.6428 & 0.3819 \\
7 & W & 33,696 & 0.6616 & 0.3837 & 23 & W & 37,848 & 0.6531 & 0.3990 \\
8 & W & 34,225 & 0.6327 & 0.3800 & 24 & P & 34,283 & 0.6360 & 0.3880 \\
9 & W & 34,184 & \textbf{0.6776} & \textbf{0.3996} & 25 & W & 34,506 & 0.6278 & 0.3704 \\
10 & W & 34,458 & 0.6421 & 0.3960 & 26 & W & 33,827 & 0.6557 & 0.3774 \\
11 & W & 34,400 & 0.6331 & 0.3619 & 27 & W & 33,557 & 0.6534 & 0.3910 \\
12 & W & 33,397 & 0.6218 & 0.3788 & 28 & W & 37,704 & 0.6149 & 0.3684 \\
13 & W & 34,383 & 0.6122 & 0.3524 & 29 & W & 37,911 & 0.6443 & 0.3492 \\
14 & W & 34,326 & 0.6523 & 0.3665 & 30 & W & 37,897 & 0.6036 & 0.3712 \\
15 & W & 34,184 & 0.6623 & 0.3750 & 31 & W & 37,811 & 0.6487 & 0.3950 \\
16 & W & 34,048 & 0.6415 & 0.3789 & 32 & W & 33,985 & 0.6354 & 0.3855 \\
        \bottomrule
    \end{tabular}
\end{table}

\subsection{Condensed trace}
\label{app:prelim:trace}
\Cref{fig:app:prelim:trace} condenses the run into the rounds that carry something the main text relies on.
Two of them are worth naming here.
Round 15 is the only round whose recipe matches an earlier one exactly, and the agent did not set out to repeat anything: it was moving the seed forward from round 14 and landed back on round 9's configuration by accident, which is what makes the pair usable as a measure of how much a repeated round moves.
Rounds 19 and 24 are the two places where what the agent says it is doing and what its plan does come apart, which is why we label actions from the difference between consecutive recipes rather than from the agent's description of a round.

\begin{figure}[t]
\centering
{\small\textbf{A single agent researches SFT data for Qwen3-1.7B-Base on IFEval and IFBench}}
\begin{tracebox}[xleftmargin=0mm,autogobble,breaklines=true,escapeinside=||]{text}[breakable=false]

|\textbf{[R1--R8: the opening move, and every move after it]}|
Agent: "It will be a single `reweight` plan with a dense, explicit
       weight on every fixed category."
-> 34,860 rows            IFEval 65.50   IFBench 37.28
|\textcolor{tracenote}{-- context compacted (R2, R6) --}|

|\textbf{[R9: the best round of the run]}|
Agent: "I'm writing round 9 as a tighter reweight, concentrating on
       the overlap between IFEval's `keywords` and IFBench's `words`."
-> 34,184 rows, seed 50   IFEval 67.76   IFBench 39.96
|\textcolor{tracenote}{-- context compacted (R9) --}|

|\textbf{[R15: a seed sweep lands back on round 9]}|
Agent: "The only changes are the round id and seed; the action type
       and dense weights stay fixed."
-> identical to R9        |\textcolor{tracewarn}{IFEval 66.23 (-1.53)   IFBench 37.49 (-2.46)}|

|\textbf{[R19--R23: the plan does not do what the round says]}|
Agent: "... an exact return to the best historical weight profile
       and a fresh seed."
-> |\textcolor{tracewarn}{the plan uses round 9's realized row counts as weights}|
   capital_word_frequency: 25 -> 157

|\textbf{[R24: the one round that is not a reweight]}|
Agent: "... exact-row `replay` of round 9, using round 9's original
       dense quota vector and row manifest."
-> |\textcolor{tracewarn}{34,283 rows: 599 of round 9's rows missing, 698 added}|

|\textbf{[R30--R33: still the same move]}|
Agent: "I'm tightening the support set now: dropping the lower-value
       families and pushing more mass into the benchmark-relevant ones."
Agent: "I'm keeping the best-known weight vector and moving the seed
       forward again."
|\textcolor{tracenote}{-- context compacted, ten times in all --}|

|\textbf{Never in 33 rounds: a row edited, a response replaced, an example}|
|\textbf{written. Never once: a mention of time.}|
\end{tracebox}
\caption{Condensed trace of the single-agent run, 33 rounds over 11.5h.
Red marks an outcome that contradicts the round's own description. Rounds not
shown repeat the pattern of others.}
\label{fig:app:prelim:trace}
\end{figure}

\subsection{What fills the context}
\label{app:prelim:context}
The deciding agent runs as one session of 877 model calls over 690.9 minutes, reaching 94.74\% of its 258{,}400-token window before the tenth compaction.
\Cref{tab:app:prelim:context} breaks the trace down by what produced it.

The breakdown is coarse in one respect worth stating.
Categories are assigned from the command that produced each record, so evaluation output that the agent reads through a general file read is counted as a file read rather than as evaluation output.
The shares establish that a few large payloads dominate.
They do not establish that context pressure caused the scores to stop improving.

\begin{table}[!htbp]
    \caption{Share of the accumulated trace by source.}
    \label{tab:app:prelim:context}
    \centering
    \begin{tabular*}{0.72\linewidth}{l@{\extracolsep{\fill}}rr}
        \toprule
        \textbf{Source} & \textbf{Bytes} & \textbf{Share} \\
        \midrule
        Agent reasoning              & 1{,}610{,}649 & 16.3\% \\
        Training output              & 1{,}322{,}501 & 13.4\% \\
        Re-reading the run record    & 1{,}167{,}202 & 11.8\% \\
        Applying the data action     &   953{,}273 & \phantom{0}9.7\% \\
        Tool-call records            &   903{,}931 & \phantom{0}9.2\% \\
        Other file reads             &   889{,}396 & \phantom{0}9.0\% \\
        Evaluation output            &   881{,}025 & \phantom{0}8.9\% \\
        Compaction summaries         &   820{,}327 & \phantom{0}8.3\% \\
        Workspace search             &   404{,}512 & \phantom{0}4.1\% \\
        Assistant messages           &   359{,}744 & \phantom{0}3.6\% \\
        Job submission               &   259{,}481 & \phantom{0}2.6\% \\
        Polling job status           &   169{,}621 & \phantom{0}1.7\% \\
        Data index inspection        &   113{,}706 & \phantom{0}1.2\% \\
        \midrule
        \textbf{Total}               & \textbf{9{,}855{,}368} & \textbf{100\%} \\
        \bottomrule
    \end{tabular*}
\end{table}

\section{Details of AutoSFT implementation}
\label{app:method}

\subsection{Capability taxonomy}
\label{app:method:taxonomy}
Deciding what data a round needs is easier when the benchmark already says what it is testing.
Instruction following does say so.
Its prompts carry constraints of known kinds, and a benchmark that groups them lets a weak group be read directly off the scores, after which the agent can raise that group's share, label existing rows by the capability they teach, or ask for new rows aimed at it.
Most domains do not come with such a grouping, and we did not want a system that only works where one exists.
The meta agent therefore builds and maintains the grouping itself.
It starts from whatever structure the benchmark provides, which for instruction following is the published constraint types, and extends it as rounds expose distinctions the original grouping did not make.
When a benchmark provides nothing, the agent derives the groups from the failures it observes.
Every leaf of the resulting taxonomy is a requirement a program can decide, such as answering without using a comma, and carries the program that decides it.
Rows in the pool are labelled by the leaves they satisfy, which is what allows the meta agent to select data by capability rather than by source and gives the synthesis subagent stable names to write against.
The taxonomy is rebuilt for a new domain, and \Cref{sec:generalization} reports how far that rebuilding got.

\subsection{Memory}
\label{app:method:memory}
The history $H_{<t}$ of \Cref{sec:setup:loop} is the only thing a from-base loop carries between rounds, and AutoSFT puts more in it than a list of recipes and scores.
Alongside those, it records the hypothesis a round registered before training, the verdict that hypothesis received, the lesson the verdict left, and the proposals the meta agent read but did not take.
Each entry is keyed by round, so a later round can read what an earlier one expected as well as what it got.

Registering the hypothesis first is what makes it worth recording.
A round states the data change it plans, why that change should move the metric, the round its result will be compared against, the direction it expects, and the finer observation that would show the change working as intended.
The submission path refuses a round whose hypothesis was never written down.
A prediction written after the results are known can always be adjusted to fit them, which is what the ordering prevents.

A verdict is read on two separate questions.
The first asks what the benchmark mean did against the comparison round, judged only as a direction and only against the variation measured between repeated rounds.
The second asks whether the finer observation actually appeared in the evaluation outputs.
Keeping them apart matters because they can disagree, and a disagreement is informative.
A score that rose without its intended effect appearing has some other cause, and an effect that appeared without a score change means the slice it touched was too small or a loss elsewhere cancelled it.

What a verdict leaves behind is a short statement of what it means for later rounds.
A later round may disagree with an earlier statement by saying so in its own entry.
Nothing already written is edited: a registered prediction cannot be changed and a decided round is not re-judged, so the memory holds what the run believed at the time rather than a version tidied up afterwards.

\subsection{Skill library}
\label{app:method:skills}
A skill is a written procedure owned by one role.
Skills do not call one another, and the sequence belongs to the agent that runs them.
\Cref{tab:app:skills} lists the library by role.

\begin{table}[t]
    \caption{The skill library, by the role that owns each skill.}
    \label{tab:app:skills}
    \centering
    \small
    \setlength{\tabcolsep}{5pt}
    \begin{tabular*}{\linewidth}{l@{\extracolsep{\fill}}lp{0.54\linewidth}}
        \toprule
        \textbf{Role} & \textbf{Skill} & \textbf{Purpose} \\
        \midrule
        \textbf{Meta agent}
        & recipe authoring
        & \raggedright Assemble the round's training set from the pool under the token budget. \tabularnewline
                & synthesis handbook
        & \raggedright Describe the role of synthesis, the structure and update rules of its pool, and how to use its outputs. \tabularnewline
        \midrule
        \textbf{Executor}
        & training submission
        & \raggedright Submit a training job to the cluster. \tabularnewline
        & job lifecycle
        & \raggedright Monitor, collect, and stop cluster jobs. \tabularnewline
        & checkpoint conversion
        & \raggedright Convert a trained checkpoint for evaluation. \tabularnewline
        & evaluation submission
        & \raggedright Submit one evaluation job per benchmark. \tabularnewline
        \midrule
        \textbf{Analysis}
        & training-data examination
        & \raggedright Quality, coverage, redundancy, and repairable rows. \tabularnewline
        & failure-case analysis
        & \raggedright Passing against failing behaviour, and the supervision that covers it. \tabularnewline
        & learning-dynamics analysis
        & \raggedright Training signals read against the data that produced them. \tabularnewline
        & history analysis
        & \raggedright Past interventions, and ideas never properly tested. \tabularnewline
        \midrule
        \textbf{Synthesis}
        & row analysis
        & \raggedright Re-derivable judgements over individual rows. \tabularnewline
        & pool rewrite
        & \raggedright Isolated, source-grounded repair trials. \tabularnewline
        & batch inference
        & \raggedright Batched model calls underlying the two above. \tabularnewline
        \bottomrule
    \end{tabular*}
\end{table}

The four analysis skills are the reason four subagents disagree productively.
They share their proposal section word for word, so all four return the same kind of object, and they differ only in the evidence each is told to start from.
Diversity comes from the starting point rather than from handing each subagent a different menu of actions, which is consistent with none of them being shown the action space of \Cref{sec:setup:actions}.
Figures~\ref{fig:app:skill:exec} to~\ref{fig:app:skill:syn} show one skill from each role.

\begin{figure}[t]
\centering
{\small\textbf{Executor: evaluation submission}}
\begin{skillbox}[xleftmargin=0mm,breaklines=true,escapeinside=||]{text}[breakable=false]

|\textbf{\# Scope}|

The active run contract supplies the model/checkpoint and benchmark list, and
the eval run name is the round being evaluated. |\textcolor{skillkey}{Do not select the model,}|
|\textcolor{skillkey}{alter the benchmark set, or inspect result values.}|

|\textbf{\# Submit a converted model}|

This is the submission path the executor handoff drives. The handoff reaches
`eval_submit` only after `checkpoint_conversion` has completed, so the model
is already converted; submit it directly rather than re-running conversion.

  python evaluate.py --model-path <contract-model-path> \
    --benchmark <contract-benchmarks> --run-name round_<n> --repeats <n>

`--run-name` is required and must be the canonical `round_<n>` of the round
being evaluated, matching the training run name. An eval run whose directory
does not identify its round cannot be joined back to that round when reading
results.
\end{skillbox}
\caption{The executor is told what to run and is denied the one thing that
would let it influence the round, which is reading the scores.}
\label{fig:app:skill:exec}
\end{figure}

\begin{figure}[t]
\centering
{\small\textbf{Analysis: training-data examination}}
\begin{skillbox}[xleftmargin=0mm,breaklines=true,escapeinside=||]{text}[breakable=false]

|\textbf{\# Scope}|

Start from the pinned input snapshot named by the task context. Examine the
staged pool and, when available, the incumbent round data and its exposure.
Assess current content rather than trusting filenames, old labels, or prior
quality annotations. Look for useful material lost by selection, duplicated
or conflicting supervision, insufficient coverage, and rows whose
substantive task can be recovered by repairing the instruction, response,
or metadata.

|\textbf{\# Proposal}|            |\textcolor{skillkey}{<- identical in all four analysis skills}|

Deliver an executable full-scale plan, not a small training dataset. Name the
source populations, selection or transformation steps, quality checks and
unknowns, expected coverage, attainable token volume, budget tradeoffs, and
how the meta agent can assemble the formal set. Distinguish measured findings
from estimates and expected effects. A useful negative result or unresolved
uncertainty is valid. Do not invent a gain or apply an arbitrary quality gate.

|\textcolor{skillkey}{Analysis subagents cannot edit the shared pool, the round data, the recipe,}|
|\textcolor{skillkey}{or the memory.}| Persist the proposal through the task report and cite the
source snapshots and artifacts it rests on.
\end{skillbox}
\caption{One of the four analysis skills. The others replace only the
\emph{Scope} section, so the four subagents start from different evidence and
return the same kind of object.}
\label{fig:app:skill:analysis}
\end{figure}

\begin{figure}[t]
\centering
{\small\textbf{Synthesis worker: how the subagent generates and checks a batch}}

\begin{skillbox}[xleftmargin=0mm,breaklines=true,breakindent=0pt,escapeinside=||]{text}[breakable=false]

|\textbf{\# Research}|

Read the target's constraints and the quoted failures that motivated it, and
study the pool rows that already satisfy adjacent capabilities. Decide the
generation approach yourself; the target states what to teach and how it is
accepted, not how to produce it.

|\textbf{\# Generation}|

Design a system prompt and few-shot demonstrations for the generation model,
drawn only from pool rows or from your own candidates that already passed
validation this round. |\textcolor{skillkey}{Over-generate several times the required rows and discard every candidate that fails validation}|; when a batch
is weak, diagnose why, revise the prompt, and call again rather than lower the
standard. Rows must come from model API calls, not from a script that assembles
responses to satisfy the verifier.

|\textbf{\# Verification}|

Translate the target's requirement into acceptance criteria and build a verifier
for it. Programmatically checkable constraints are gated in full by that
verifier; semantic fit gets one model review on the final candidate set, and
read-only subagents that took no part in generation spot-check quality,
diversity, and fit. Close at least one loop of |\textcolor{skillkey}{generate $\rightarrow$ verify $\rightarrow$ review $\rightarrow$ revise}| before
delivering.

|\textbf{\# Decontamination}|

Phrase the capability requirement without reproducing evaluation wording. |\textcolor{skillkey}{An $n$-gram overlap check against the evaluation sets runs before delivery}|, so a batch teaches the capability rather than the benchmark.
\end{skillbox}
\caption{One skill given to the synthesis subagent. It fixes the internal loop
by which a worker turns a target into a delivered batch, while leaving the
generation strategy to the worker. The generation, verification, and
decontamination checks it enforces are detailed in
Appendix~\ref{app:method:synthesis-validation}.}
\label{fig:app:skill:syn}
\end{figure}

\subsection{Synthesis constraints and verification}
\label{app:method:synthesis-validation}

\textbf{Generation constraints.}
The subagent phrases each capability requirement without reproducing evaluation wording, and the generated rows pass an $n$-gram overlap check against the evaluation sets before delivery, so a batch teaches the capability rather than the benchmark.
Rows must come from model API calls rather than from a script, which prevents a worker from assembling responses that satisfy the verifier without genuinely demonstrating the capability.

\textbf{Internal verification.}
The subagent attributes each evaluation subset's requirements to an underlying capability, translates them into acceptance criteria, and builds a verifier for those criteria.
It then generates candidates, checks them against the verifier, and revises until they meet the criteria, maintaining both the criteria and the verifier as it goes so that omissions are corrected without changing the intended capability.
\section{Details of the experiments}
\label{app:exp}

\subsection{Analysis subagents}
\label{app:exp:analysis}

\Cref{fig:app:analysis-caseB} shows a case in which both runs address persistent repetition errors.
The run with analysis varies the wording and order of the cue, while the base run increases exposure to the same template.
The first raises strict repetition accuracy by 34.76 points, the second by 1.22.

\begin{figure}[t]
\begin{casebox}

\medskip
\noindent\begin{minipage}[t]{0.475\linewidth}
\raggedright
{\color{runplus}\textbf{With analysis}\quad R10 $\to$ R14}

\smallskip
The failure reading traces the errors to 792 examples that share one template, and R11--R13 substitutions do not surpass R10.

\smallskip
The meta agent keeps 88.46\% of R10's tokens and replaces the rest with 2{,}752 examples that vary cue wording and order across all 16 cells.

\smallskip
Strict repetition $39.02 \to 73.78$ \textbf{(+34.76)}.
\end{minipage}\hfill
\begin{minipage}[t]{0.475\linewidth}
\raggedright
{\color{runbase}\textbf{Base}\quad R18 $\to$ R19}

\smallskip
The ``repeat first'' directive at R18 improves repetition without raising the best score, which reads as useful but insufficient exposure.

\smallskip
The meta agent keeps the directive and expands from 500 to all 974 available examples of the same template.

\smallskip
Strict repetition $23.78 \to 25.00$ \textbf{(+1.22)}.
\end{minipage}
\end{casebox}
\caption{The second decision of \Cref{sec:exp:ablations}. Explanations reflect the reasoning recorded at the time, and strict repetition is measured per instruction.}
\label{fig:app:analysis-caseB}
\end{figure}

\Cref{tab:app:newbest} groups the data actions into the families of \Cref{fig:setup:actions} and shows the fraction of the rounds that used it which set a new best score, where a new best is a score above every earlier round in the same run.
A round can use more than one action, so it counts towards each action it used.
The base run reaches only the \emph{mixture} family, so its entries appear there alone, and the other families were never used without analysis, as in the single agent of \Cref{sec:prelim:loop}.

\begin{table}[t]
    \caption{New-best rate by data action, as the number of rounds that set a new best over the number of rounds that used the action. The run with analysis covers 30 rounds and the base run its own rounds. A dash marks a family the base run did not use.}
    \label{tab:app:newbest}
    \centering
    \small
    \begin{tabular}{lr@{\ }lr@{\ }l}
        \toprule
        \textbf{Data action} & \multicolumn{2}{c}{\textbf{With analysis}} & \multicolumn{2}{c}{\textbf{Base}} \\
        \midrule
        \multicolumn{5}{l}{\emph{Mixture}} \\
        \quad Constraint type          & 33.3\% & (10/30) & 17.2\% & (5/29) \\
        \quad Number of constraints    & 34.8\% & (8/23)  & 14.3\% & (1/7) \\
        \quad Rare constraint types    & 16.7\% & (1/6)   & 33.3\% & (2/6) \\
        \quad Language                 & 25.0\% & (4/16)  & 0.0\%  & (0/2) \\
        \quad Pool-wide selection      & 50.0\% & (2/4)   & 9.1\%  & (1/11) \\
        \quad Restore earlier data     & 22.2\% & (2/9)   & 25.0\% & (2/8) \\
        \quad Full dataset reselection & 50.0\% & (1/2)   & 20.0\% & (2/10) \\
        \midrule
        \multicolumn{5}{l}{\emph{Pruning}} \\
        \quad Validation results       & 41.7\% & (10/24) & \multicolumn{2}{c}{--} \\
        \quad Official checks          & 35.0\% & (7/20)  & \multicolumn{2}{c}{--} \\
        \quad Text length              & 42.9\% & (6/14)  & \multicolumn{2}{c}{--} \\
        \midrule
        \multicolumn{5}{l}{\emph{Repair}} \\
        \quad Edit instructions and answers & 42.9\% & (6/14) & \multicolumn{2}{c}{--} \\
        \midrule
        \multicolumn{5}{l}{\emph{Resupervision}} \\
        \quad Replace answers only     & 0.0\%  & (0/3)   & \multicolumn{2}{c}{--} \\
        \bottomrule
    \end{tabular}
\end{table}

\subsection{Synthesis subagent}
\label{app:synthesis}

\begin{figure}[!htbp]
\centering
\includegraphics[width=\linewidth]{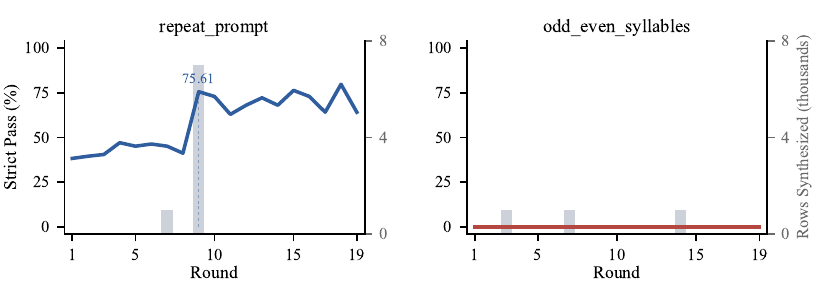}
\caption{Two capabilities over the run with synthesis. The line is the capability's whole-question strict pass rate each round, and the bars are the rows synthesized for it that round.}
\label{fig:app:synthesis:cap}
\end{figure}

\Cref{fig:app:synthesis:cap} follows the two capabilities named in \Cref{sec:exp:synthesis} over the full run with synthesis.
For \texttt{repeat\_prompt}, the line shows the capability's strict pass rate staying near 40\% through the first eight rounds, then rising to 75.6\% at round 9, the round its batch enters, and remaining between 62\% and 80\% for the rest of the run.
For \texttt{odd\_even\_syllables}, the pass rate stays at zero despite batches in three separate rounds.
The bars give the number of rows synthesized for the capability.

\begin{figure}[!htbp]
\centering
\includegraphics[width=\linewidth]{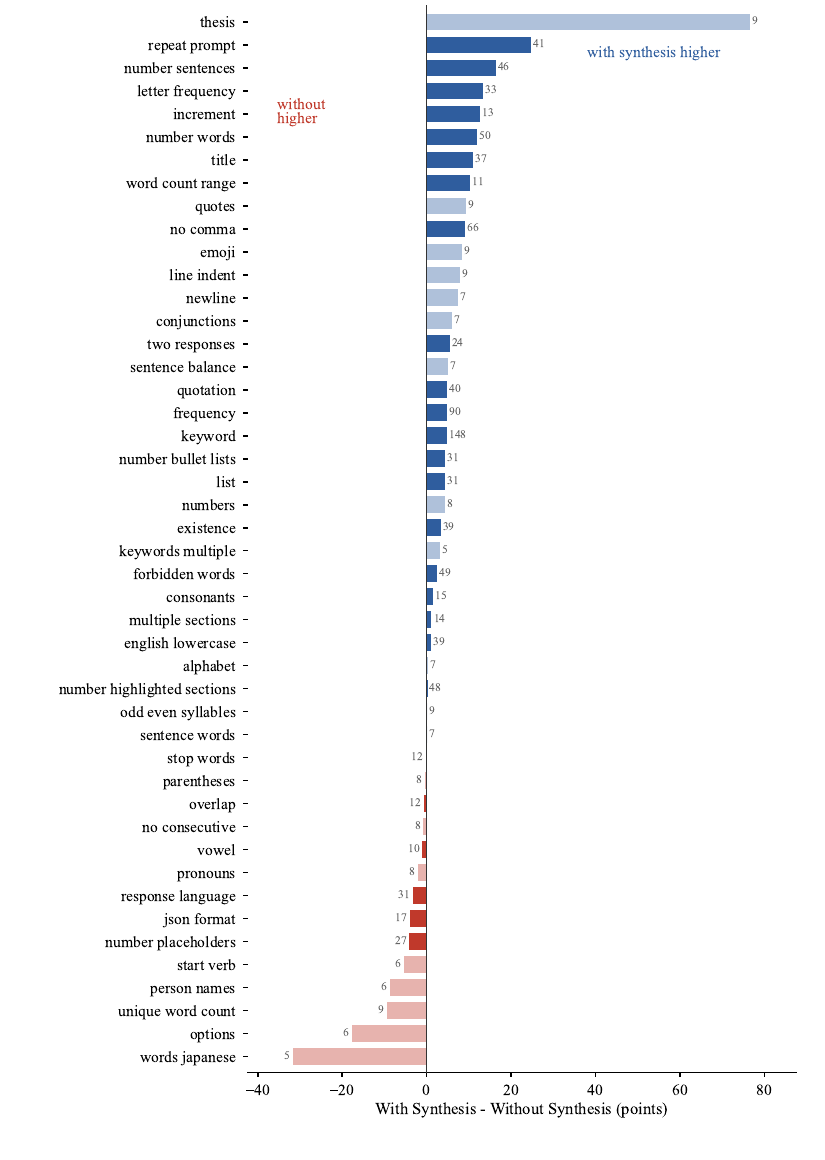}
\caption{Per-capability comparison of the two runs of \Cref{sec:exp:synthesis}.}
\label{fig:app:synthesis:advantage}
\end{figure}

\Cref{fig:app:synthesis:advantage} widens this to every capability the synthesis subagent targeted.
For every capability targeted by the synthesis subagent, each bar shows the difference in whole-question strict pass rate between the run with synthesis and the run without synthesis, averaged over the last five rounds of each run.
A question passes only when all of its co-occurring requirements pass. The number beside each bar is the number of questions the capability contains, and capabilities backed by fewer than ten questions are drawn faded because their differences are likely to reflect small-sample noise.
Most targeted capabilities show higher rates with synthesis, and the advantage is largest and best supported for repetition, sentence and word counts, and comma avoidance, each backed by dozens of questions.
A few capabilities show lower rates, most of which are backed by only a handful of questions.

\noindent\textbf{Target selection checks data supply rather than relying on aggregate capability scores alone.}
The subagent selects a target when the pool contains far fewer than the 1000 rows it treats as sufficient, and it records the count it finds.
For example, the R9 targets matched only 4 to 161 staged rows each. The subagent excluded capabilities that already had ample data, reasoning that generating new examples cannot help a capability the pool already covers.
A capability with a high aggregate score can therefore still be a target, because what is scarce is a specific constraint combination rather than the capability as a whole. An example is repetition together with a number word, as opposed to number words on their own.
The score of an exact combination cannot be measured directly, because the benchmark contains at most one question for it. The comparison in \Cref{fig:app:synthesis:advantage} is therefore made at the aggregate capability level.

\section{Experimental Details and Resource Usage}
\label{app:implementation-cost}

\begin{table}[t]
\centering
\caption{Settings shared by every experiment.}
\label{tab:app:shared}
\begin{tabular}{ll}
\toprule
\textbf{Component} & \textbf{Setting} \\
\midrule
\multicolumn{2}{l}{\emph{Agent}} \\
Harness & OpenAI Codex \\
Research agent & GPT-5.5, medium reasoning effort \\
Generation backend & DeepSeek V4 Flash \\
\midrule
\multicolumn{2}{l}{\emph{Training}} \\
Base model & Qwen3-1.7B-Base \\
Framework & Megatron-LM \\
Hardware & $8\times$ NVIDIA H800 \\
Epochs per round & 3 \\
Batch size (global / micro) & 16 / 1 \\
Optimizer & Adam ($\beta_1{=}0.9$, $\beta_2{=}0.95$, $\epsilon{=}10^{-8}$), weight decay $0.1$ \\
Learning rate & $10^{-4}\!\rightarrow\!10^{-7}$, cosine decay per round, 5\% warmup \\
\midrule
\multicolumn{2}{l}{\emph{Evaluation}} \\
Framework & OpenCompass with a vLLM backend \\
Hardware & $1\times$ NVIDIA H800 \\
Decoding & temperature $0.7$, zero-shot prompting \\
\bottomrule
\end{tabular}
\end{table}

All experiments share one implementation for agent execution, supervised fine-tuning, and evaluation.
A research agent coordinates each round and makes the data decisions, while a separate batch-inference backend generates and reviews synthesized examples.
Every candidate recipe is fine-tuned from the same base checkpoint, and only the research history and the evaluation feedback carry across rounds.
\Cref{tab:app:shared} lists the settings common to every run.

The comparisons differ in only a few settings, listed in \Cref{tab:app:percomparison}.
The analysis and synthesis ablations draw from the same open instruction-following pool of \Cref{sec:prelim} but under different token budgets, and the code experiments use a Python pool with a shorter sequence length.
A benchmark score is reported as \emph{avg@k}, the mean over $k$ stochastic evaluations of one checkpoint. IFEval and IFBench use $k{=}10$ and MBPP+ uses $k{=}3$.

\begin{table}[t]
\centering
\caption{Settings that differ across the three comparisons. Within each comparison the run with and without the component share these settings, so a difference is attributable to the component alone.}
\label{tab:app:percomparison}
\begin{tabular}{lccc}
\toprule
 & \textbf{Analysis} & \textbf{Synthesis} & \textbf{Code} \\
\midrule
Initial pool & \multicolumn{2}{c}{Open ifeval-like pool (\Cref{sec:prelim})} & 742{,}842 Python pairs \\
Token budget & 3M & 100M & 100M \\
Sequence length & 32{,}768 & 32{,}768 & 8{,}192 \\
Benchmarks & IFEval & IFEval, IFBench & MBPP+ \\
Evaluation & avg@10 & avg@10 & avg@3 \\
\bottomrule
\end{tabular}
\end{table}

\noindent\textbf{Resource usage.}
We estimate the cost of a research run from the synthesis experiment, which uses the largest per-round budget of 100M tokens.
Each round prepares data under agent guidance, fine-tunes one recipe, and evaluates the resulting checkpoint.
Training takes about $0.8$ GPU-hours per round on 8 H800 GPUs and evaluation about $0.15$ GPU-hours on one, so a 20-round run costs roughly 130 GPU-hours.
The research agent and the generation backend together consume about $2.6\times10^{8}$ tokens per round, counting input, output, and reasoning tokens, for an estimated API cost near USD~95 per round, or USD~1{,}900 over the run.
The paper reports about ten distinct runs and a few hundred rounds in total, counting a shared baseline once.
Scaling each experiment by its token budget gives an order-of-magnitude total of $10^{3}$ GPU-hours and USD~$10^{4}$, with real usage varying by hardware utilization, evaluation protocol, and agent activity.

\section{Limitations}
\label{app:limitation}
All results are obtained on one base model at one scale (Qwen3-1.7B-Base), so the effects we report may not hold at other scales, and a capability this model cannot learn from more data may be learnable by a larger one.
Each comparison runs one search per configuration, and although repeated evaluation bounds the noise of a single checkpoint, it does not bound the variation between independent searches.
Synthesis raises the aggregate score through a small number of learnable capabilities that the pool undersupplies, and it does not move capabilities the model cannot yet follow.
The generalization study covers a single new domain with short runs, so it demonstrates that the loop can transfer but does not establish how broadly.

\end{document}